%% file: main.tex
\documentclass[runningheads,table,xcdraw]{llncs}
\usepackage{pgfplots} 
\usepackage{graphicx}
\usepackage{subfiles}
\usepackage{tabularx}
\usepackage{xcolor}
\usepackage{float}
\usepackage{svg}
\usepackage{siunitx}
\usepackage[colorinlistoftodos,prependcaption,textsize=tiny]{todonotes}

\begin{document}
\title{Information Extraction from Documents: Question Answering vs Token Classification in real-world setups}
\titlerunning{Information Extraction from Documents: QA vs TC in real-world setups}
%
\author{Laurent Lam,  Pirashanth Ratnamogan, Joël Tang, William Vanhuffel, Fabien Caspani}
%
\authorrunning{L. Lam et al.}
%
\institute{BNP Paribas 
(laurent.lam;pirashanth.ratnamogan;joel.tang)@bnpparibas.com
(william.vanhuffel;fabien.caspani)@bnpparibas.com
}
\maketitle              
\begin{abstract}
\subfile{Sections/1-Abstract}
\end{abstract}
\section{Introduction}
\subfile{Sections/2-Introduction}

\section{Problem definition}
\subfile{Sections/3-ProblemDefinition}

\section{Related work}
\subfile{Sections/4-RelatedWork}

\section{Experimental setups}
\subfile{Sections/5-TaskDescription}

\section{Datasets}
\subfile{Sections/6-Datasets}

\section{Method \& Experiments}
\subfile{Sections/7-ApproachExperiments}

\section{Conclusion}
\subfile{Sections/8-Conclusion}


\bibliography{main}
\bibliographystyle{splncs04}

\end{document}

%% file: Sections/1-Abstract.tex
Research in Document Intelligence and especially in Document Key Information Extraction (DocKIE) has been mainly solved as Token Classification problem. Recent breakthroughs in both natural language processing (NLP) and computer vision helped building document-focused pre-training methods, leveraging a multimodal understanding of the document text, layout and image modalities. \\
However, these breakthroughs also led to the emergence of a new DocKIE subtask of extractive document Question Answering (DocQA), as part of the Machine Reading Comprehension (MRC) research field. \\
In this work, we compare the Question Answering approach with the classical token classification approach for document key information extraction. We designed experiments to benchmark five different experimental setups : raw performances, robustness to noisy environment, capacity to extract long entities, fine-tuning speed on Few-Shot Learning and finally Zero-Shot Learning. \\
Our research showed that when dealing with clean and relatively short entities, it is still best to use token classification-based approach, while the QA approach could be a good alternative for noisy environment or long entities use-cases.\\

\keywords{Document Key-Information Extraction \and Machine Reading Comprehension \and Named Entity Recognition \and Token Classification\and Document Question Answering.}

%% file: Sections/2-Introduction.tex
Document understanding is a key research area with a growing industrial interest. 
Many businesses manually process thousands of documents for recurrent tasks.
As part of document understanding, information extraction is a complex task, targeting to extract key structured information from unstructured documents.
It is an important but complex process as documents can take several shapes (contracts, invoices, reports ...) with their various inherent challenges (long documents, complex layouts, tables ...)

Recently, multi-modal approaches combining natural language processing, computer vision and layout understanding proved to be effective in this configuration.
In the literature two standard approaches are used. The classical approach formulates the information extraction task as a token classification approach (i.e. we classify each token to belong to a specific entity or not). The second formulates the information extraction task as a span extraction task (i.e. we search for the beginning and end of a given entity), often represented as a question - answer with fixed questions.

Our study is the first one undertaking an empirical study comparing the two approaches in complex scenarios using multiple datasets. 
It relies on LayoutLM \cite{layoutlm}, a standard backbone commonly used in the document understanding tasks.

Our contributions are as follows:
\begin{itemize}
    \item We propose multiple scenarios in order to emulate real-world complexity on open source datasets,
    \item We pursue the first study comparing token classification and question answering approaches in the information extraction task,
    \item We state in which setting to use one approach or the other one.
\end{itemize}

%% file: Sections/3-ProblemDefinition.tex
\subsection{Information Extraction from Documents} 

Information Extraction is a sub-task of document understanding that aims at extracting structured information from unstructured data. 

Traditionally, extracting information from documents consisted in classifying each token of the text as belonging to a certain class (one per attribute/entity). The IOB (Inside-Outside-Beginning) tagging \cite{iobtagging} introduced a B(eggining) token class declaring the start of a new entity. 

Several benchmark datasets are widely used, such as FUNSD, CORD or SROIE \cite{sroie,funsd,cord}. We note that existing benchmarks are heavily biased towards short entities. For instance, clauses to be extracted from long documents are not covered by these datasets.

In our study, we will explore two different approaches to perform such Information Extraction:
\begin{itemize}
    \item via classical Token Classification,
    \item via a Question Answering approach.
\end{itemize}

\subsection{Machine Reading Comprehension}

Machine reading comprehension is an active research field, belonging to the Natural Language Processing (NLP) field \cite{zheng202mrcsurvey} and overlaps with the Question Answering task.

The QA task is a NLP task and consists in processing a given question and outputting the answer. Both the question and the answer are in natural language. More specifically, the question is free-text (prompt) and the answer is extracted as one or multiple spans from a given text (the \textit{context}) since we consider the \textbf{Extractive QA} setup.
In that case, the concatenation of the question and the context forms the input of the model and the output is one or multiple contiguous spans of the text (denoted by start and end token indexes).\\

%% file: Sections/4-RelatedWork.tex
The Information Extraction problem was first turned into a sequence labeling task which is a conceptually simple approach and led to good results but it does not allow models to separate two consecutive entities of the same class. 

When the IOB tagging \cite{iobtagging} was introduced, it led to more accurate tagging and many variants of tagging schemes exist with variable performances over the models and the dataset typology \cite{tagging_scheme_benchmark}.

Recent breakthroughs in both natural language rocessing and Computer Vision led to notable improvements in the key-information extraction (KIE) task with the emergence of various efficient document pre-training methods such as the Transformer-based LayoutLM family of models \cite{layoutlmv3,layoutlmv2,layoutlm}. These models introduced a multimodal pre-training approach, standing out not only due to their difference in architecture but mostly from their incorporation of text, layout and image modalities into their pre-training for document image understanding and information extraction tasks. It leverages both text and layout features and incorporates them into a single framework which is why this family of models or variants are used by many research works. \cite{douzon:hal-03676134,DBLP:journals/corr/abs-2111-04045,DBLP:journals/corr/abs-2109-00442}

However the data quality remains essential and noisy datasets, missing tags or errors in labels are common in various real-life use-cases. Research work have been performed to detect or learn from such noisy settings \cite{https://doi.org/10.48550/arxiv.2210.03920,zhou-chen-2021-learning} in terms of Token Classification. 
\\

Another approach could also be considered in order to perform Information Extraction.

The machine reading comprehension field has also known many breakthroughs \cite{zheng202mrcsurvey}, in particular for the question answering task. At the intersection of information extraction and question answering, a few research work focused on reframing the classical token classification problem as a MRC one \cite{https://doi.org/10.48550/arxiv.1910.11476}.

Some propose to convert each attribute into a question, and to identify the answer span corresponding to the attribute value in the context \cite{mengge-etal-2020-coarse,10.1145/3394486.3403047}. This idea was followed by several incremental improvements, such as asking multiple questions in a single pass \cite{shrimal-etal-2022-ner}.

We note the scarcity of question answering datasets designed for data and questions typically framed as long document information extraction problems. For instance, SQuAD \cite{rajpurkar-etal-2016-squad}, SQuADv2 \cite{rajpurkar-etal-2018-know} and Natural Questions \cite{kwiatkowski-etal-2019-natural} all propose questions related to text comprehension, but a few include classes to be directly extracted from the text. The CUAD dataset \cite{hendrycks2021cuad} a comprehensive QA dataset on legal documents which are long by their nature: some questions ask to retrieve contract dates, parties names or non-competing clauses. Such entities to be extracted can be short (person names, dates) or very long (clauses to be retrieved in a long contract).

Li et al. \cite{https://doi.org/10.48550/arxiv.1910.11476} studied this particular topic by focusing on the adaptation of the specific NER into a MRC task.\\

Therefore, our research work introduces a new review of both Token Classification and machine reading comprehension approaches, beyond the NER adaptation previously presented. In such field of Document Information Retrieval, we experiment with various settings from standard benchmarks, noisy environment, long entities, to few-shot or zero-shot learning setups, an extensive comparison that has not been done so far in the literature.

All our experiments will be using a LayoutLM model backbone \cite{layoutlm} since it represents a standard baseline for Information Retrieval. It is also easy to set up towards a token classification task or a question answering task.

%% file: Sections/5-TaskDescription.tex
Open-source datasets usually contain clean labels and data which facilitates the benchmark of various models. However, in the industry, the data quality does not necessarily reach such standards and can present various difficulties for the model to actually learn to extract the correct entities.

In our work, we will focus on 4 different experimental settings.

\subsection{Noisy environments}

Annotating documents and creating training data is an expensive and essential part for building a supervised model.

As part of the information extraction task, annotating without any human errors is extremely costly. 
For example, the annotation of 510 contracts in the legal CUAD dataset is valued at \$2 million  \cite{hendrycks2021cuad}.

Hence, it is common to assume a certain level of noise in the dataset. 
For example, the original version of FUNSD contains too many annotation errors \cite{vu2020revising}.

In the information extraction task, the noise may be as follows:
\begin{itemize}
    \item some annotations may be missing,
    \item some texts may be incorrectly annotated (i.e. "Barack Obam" instead of "Barack Obama"),
    \item an entity can be partially annotated (i.e. "Obama" instead of "Barack Obama").
\end{itemize}

Therefore, it is essential to be able to create robust and performing models even if noise is present in the dataset.

\subsection{Long entities \& Long documents}

Extracting long entities represents a common challenge for information extraction and QA tasks. Often, they are found on long documents like contracts and other legal-binding documents (often composed of multiple dozens of pages). In this case, one entity frequently corresponds to multiple sentences.

The first issue is when the number of input tokens in the document exceeds the maximum capacity of the model. Transformer architectures, which are the state-of-the-art for the defined tasks, cannot exceed a fixed size (pretrained positional embeddings size is often set at 512 for common architectures such as BERT \cite{bert}). This is mainly because of their quadratic complexity with respect to the number of input tokens ($O(n^2)$). 

The common fallback is to divide the text into chunks such that they can properly fit in the model. However, this is an unsatisfying approach when answers are expected to be long entities, as the probability that the answer is overlapping on at least two chunks is significant.

Another common approach is to alleviate the memory consumption issue, often by truncating the attention matrix format, by using diverse inductive biases. BigBird \cite{bigbird} proposes a sampling methodology to choose the tokens used for self-attention at any position. Finally, general approaches to reduce memory consumption of models can be used, such as gradient checkpointing, at the expense of increased time complexity.

For Token Classification, long entities are also a challenge when defining the task. Models are typically trained to classify the tokens given the IOB tagging scheme \cite{iobtagging}. As entities become longer, this model output becomes very sensitive to errors. For example, one common issue is when the model predicts a sequence like "BIIIIOIIIII".

While the token classification task requires the entity classification for each token, the question answering task is only trying to classify a token as the start or the end of an entity. Long entities that overlap over different chunks could therefore be treated more independently with less dependency to contiguous context in QA than token classification.

\subsection{Few-Shot Learning}
Few-Shot Learning consists in feeding a machine learning model with very few training data to guide and focus its future predictions, as opposed to a classical fine-tuning which require a large amount of training labeled data samples for the pre-trained model to adapt to the desired task with accuracy.

It represents a major challenge in the industry since large dataset labelling is extremely costly. Therefore, achieving great performances with only a few labeled data samples is an essential goal to reach.

\subsection{Zero-Shot Learning}

Zero-Shot Learning is a problem setup which consists in a model learning how to perform a task it never did before, here: classify unseen classes.

%% file: Sections/6-Datasets.tex
In order to evaluate the different approaches and the various learning settings, multiple datasets are available.
Some of them allow to benchmark different Document Information Extraction tasks, and in our work we will use FUNSD \cite{funsd}, SROIE \cite{sroie}, Kleister-NDA \cite{Kleister} and CUAD \cite{hendrycks2021cuad} that are key information extraction datasets respectively from forms, receipts and contracts. We will also use an internal dataset of Trade Confirmations.

In this work, we focus on the key information extraction task as a common benchmark between visually-rich document understanding and machine reading comprehension.

\begin{table}[h]
\resizebox{\columnwidth}{!}{%
\begin{tabular}{|c|c|ccc|c|}
\hline
              & & \multicolumn{3}{c|}{\textbf{Token Classification}} & \textbf{QA} \\ \hline
              \textbf{Datasets} & \textbf{Split} & \textbf{Documents} &  \textbf{Entities}  & \textbf{Categories} & \textbf{QA Samples}\\ 
      \hline
      SROIE & train & 626 & 2655 & 4 & 2498\\
            & test & 346 & 1462 & & 1384 \\
      
      \hline
      FUNSD & train & 149 & 6536 & 3 & 440\\
       & test & 50 & 1983 &  & 117 \\
      \hline
      Kleister & train & 254 & 2861 & 4 & 744 \\
      NDA  & validation & 83 & 928 &  & 254 \\
      \hline
      CUAD & custom train & 408 & 1849 & 10 & 787\\
       & custom test & 102 & 421 &  & 168\\
      \hline
      Trade & custom train & 170 & 1970 & 12 & 1857\\
      Confirmations  & custom test & 42 & 544 &  & 465\\
      \hline
\end{tabular}%
}
\caption{Dataset statistics comparison between QA and Token Classification approaches for SROIE, FUNSD, Kleister NDA, CUAD and Trade Confirmations}
\label{tab:datasets_comparison_ner_qa}
\end{table}

\subsection{SROIE}
The Scanned Receipts OCR and key Information Extraction (SROIE) \cite{sroie} dataset was introduced at the ICDAR 2019 conference for competition. Three tasks were set up for the competition: Scanned receipt text localisation, scanned receipt optical character recognition and key information extraction from scanned receipts.
In this work, SROIE will be mainly used for the different benchmarks, with entities to extract among \textit{company}, \textit{address}, \textit{total} or \textit{date}.

\subsection{FUNSD}
The Form Understanding in Noisy Scanned Documents (FUNSD) \cite{funsd} dataset was introduced in 2019 and has been a classical benchmark in recent Document KIE research works. The different entities are labeled as \textit{header}, \textit{question} or \textit{answer}.
In this work, we will be using a revised version of this dataset \cite{revisedFunsd} with cleaned annotations.

\subsection{Kleister NDA}
The Kleister NDA \cite{Kleister} dataset is composed of long formal born-digital documents of US Nondisclosure Agreements, also known as Confidentiality Agreements, with labels such as \textit{Party}, \textit{Jurisdiction}, \textit{Effective Date} or \textit{Term}.

\subsection{CUAD}

Contract Understanding Atticus Dataset (CUAD) \cite{hendrycks2021cuad} is a dataset introduced in 2021, with  classes representing information of interest to lawyers and other legal workers when analyzing such legal documents. For instance, short labels (\textit{contract date}, \textit{parties names}, etc.) exist, just as long labels (\textit{outsourcing agreement}, \textit{outsourcing agreement}, etc.) where clauses are labeled.

\subsection{Trade Confirmations} Trade confirmations are financial documents reporting the details of a completed trade. It is well structured as the document comes from a limited number of counterparties. It is composed of one-page PDFs detailing derivatives products and 13 different entities to extract (\textit{price}, \textit{volume}, \textit{trade date} ...).

%% file: Sections/7-ApproachExperiments.tex
\subsection{Question Answering}
Extractive QA datasets are based on tuples of contexts, queries and answers ($c$, $q$, $a$). The answers can be found in the contexts and are associated with their respective start indices and lengths in each context.

In order to adapt the QA task as an Information Extraction task into an entity retrieval task, we turned the usual query into a generic question in natural language about the given label: \textbf{\textit{What is the $<$LABEL$>$?}}.\\

For each label to be found in a document, we create a QA data sample composed of the document's context, the generated query and the start indices and lengths of the answers. One document could therefore appear multiple times in the QA dataset since it would be associated with different queries and answers.

This procedure allows us to convert token classification-based datasets into question answering datasets. As shown in Tab. \ref{tab:datasets_comparison_ner_qa}, the number of samples for each dataset is significantly larger when converted to QA task due to the dependence on the number of entities and labels.

\subsection{Common setup}

In this work, as mentioned, we will base our experiment on a unique LayoutLM \cite{layoutlm} backbone and especially the style-based embeddings variant introduced by Oussaid et al. \cite{DBLP:journals/corr/abs-2111-04045} due to its efficiency and effectiveness.

\begin{figure}[h]
  \centering
  \includegraphics[width = 13cm]{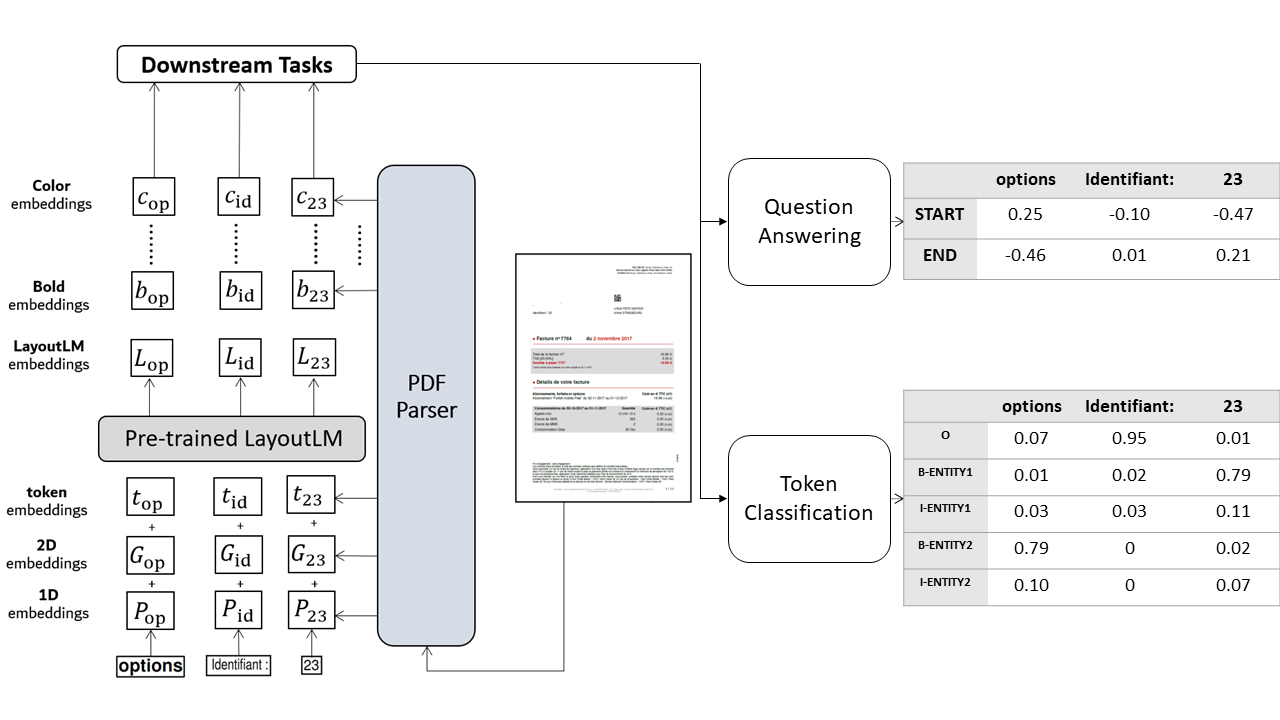}
  \caption{Processes of LayoutLM + style-based embeddings with QA and Token Classification tasks}
  \label{fig:architecture}
\end{figure}

Based on that backbone as shown on Fig. \ref{fig:architecture}, the difference lies on the last classification layer when trying to classify the different entities for each dataset in token classification whereas the last layer in QA classifies start and end.

For the token classification approach, we use a batch size of 2 and an Adam optimizer with an initial learning rate of $2*10^{-5}$.
Then throughout the training, if there is no increase in the validation F1-score after 10 epochs, the learning rate is divided by 2. We stop fine-tuning when the learning rate goes below $10^{-7}$.

Regarding the experimental setup for fine-tuning, for the QA approach, we use a batch size of 4 and an Adam optimizer with an initial learning rate of $2*10^{-5}$. The experiments also ran with gradient accumulation steps of 2. We run the fine-tuning  with early stopping on F1-score as well.\\

\subsection{Vanilla setting}

In order to compare both approaches, we started by benchmarking the performances of LayoutLM as token classification and LayoutLM as QA on the datasets.


\begin{table}[h]
\resizebox{\columnwidth}{!}{
    \begin{tabular}{|c|SSS|SSS|}
    \hline
                  & \multicolumn{3}{c|}{\textbf{LayoutLM$_{base}$ TC}} &  \multicolumn{3}{c|}{\textbf{LayoutLM$_{base}$ QA}} \\ \hline
                  \textbf{Model} & \textbf{F1}   & \textbf{Precision} & \textbf{Recall}               & \textbf{F1}   & \textbf{Precision} & \textbf{Recall}  \\ 
          \hline
          SROIE & \textbf{95.79} & 95.36 & 96.24 & 93.78 & 93.78 & 93.78\\
          \hline
          FUNSD & \textbf{86.57} & 87.23 & 86.03 & 6.84 & 55.90 & 4.03\\
          \hline
          Kleister NDA & \textbf{76.81} & 77.60 & 76.40 & 32.58 & 53.71 & 24.03\\
          \hline
          Trade Confirmations & \textbf{97.14} & 97.24 & 97.06 & 87.88 & 94.45 & 83.09\\
          \hline
    \end{tabular}
}
\caption{Performance of Token Classification and QA LayoutLM information extraction models on SROIE, FUNSD, Kleister NDA and Trade Confirmations}
\label{tab:benchmark}
\end{table}

As represented in Tab. \ref{tab:benchmark}, the question answering approach's performances are not consistent across datasets since it performed very well on SROIE whereas it achieved only poor results on Kleister NDA and did not succeed in capturing some of the labels on FUNSD dataset.
On the other hand, the token classification approach achieved from acceptable to very good results on each dataset.

The poor results from QA approach on FUNSD could be explained by the lack of specific semantics from its labels (\textit{header}, \textit{question} or \textit{answer}). Indeed this approach will first process the question query via its semantics but the \textit{question} or \textit{answer} is not necessarily a refined or accurate query for the model, especially when taken independently.

Another drawback of the QA approach that may lead to poorer performances on datasets with multiple tags per label is its multi-responses handling for a given label. Since the SQuAD v2 \cite{rajpurkar-etal-2018-know} dataset with the introduction of unanswerable questions, the model must also determine when no viable answer can be extracted from the context and should abstain from answering. But when there are multiple (\textit{k}) expected answers, taking the \textit{k}-top answer outputs while filtering out the considered non-viable answers from the model does not necessarily lead to the wanted tags. As in SQuAD v2, the sum of the logits of the \textit{start} and \textit{end} tokens must then be positive in order to be considered. 

On the internal Trade Confirmations dataset, the question answering dataset achieved correct performances with 87.88 weighted average F1-score whereas the token classification model achieved almost perfect predictions with 97.14 weighted average F1-score.

\subsection{Noisy environments}

In order to generate a noisy dataset from the SROIE dataset, we randomly sub-sampled from the tags of each document using different sub-sampling ratios. That setting allows us to recreate an environment where datasets are not fully but only partially tagged.

We decided to take as sub-sampling tag ratios \textit{10\%}, \textit{30\%}, \textit{50\%}, \textit{70\%}, \textit{90\%}. This sub-sampling procedure is applied on the training and validation sets but the test set remains the same fully annotated split. It can also happen that all tags from a document are discarded using this procedure, the document is then also discarded from the training or validation set.

We also performed the sub-sampling and training using 5 different random seeds in order to assess the stability and statistical significance of the results.

\begin{figure}[h]
  \centering
  \includegraphics[width = 12cm]{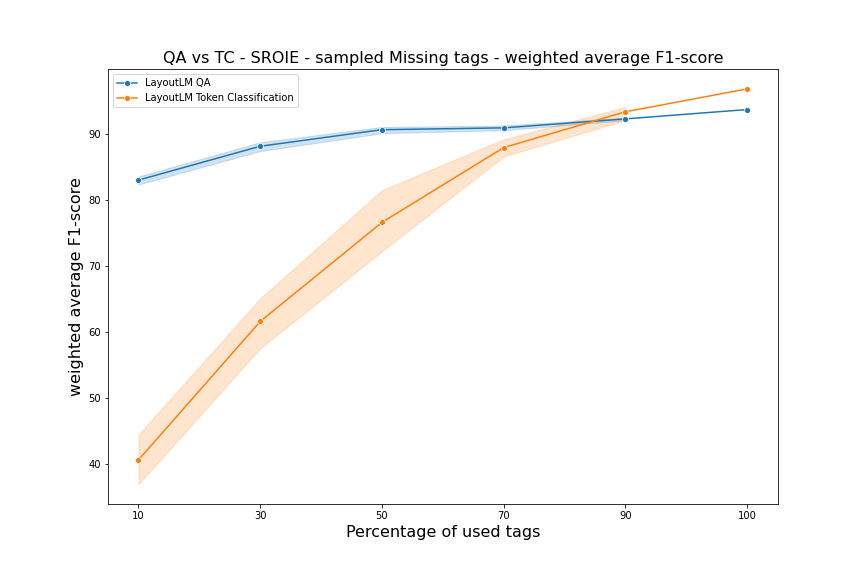}
  \caption{SROIE: QA vs Token Classification performances using partial random missing tags}
  \label{fig:sroie_missing_tags}
\end{figure}

As shown in Fig. \ref{fig:sroie_missing_tags}, we notice that the Token Classification approach is seriously affected by this noisy environment with more than 50\% of average decrease in weighted F1-score when using only 10\% of the tags (40.66\%) whereas for the QA approach, the F1-score only decreased on average of 9\% achieving on average 83.09\% of weighted F1-score.

We notice that the results from the token classification approach have a greater variance than the results from QA.

This is due to the fact that the QA approach is only fed with positive training samples compared to the token classification approach that is provided all token classifications, even where there are no tags. It therefore learns not to predict certain tokens or values.

The performances of the QA approach is therefore more consistent using different ratios of tag sub-sampling than the token classification approach.

However the token classification approach reaches slightly higher performances than the QA approach with the fully annotated dataset as stated in the Vanilla setting.

\subsection{Long entities \& Long documents}

\begin{table}[h]

\begin{tabular}{|c|c|c|c|}
\hline
              \textbf{Label} & \textbf{Count} & \textbf{Average \#characters} & \textbf{Median \#characters} \\ 
      \hline
      \hline
      Affiliated License Licensor & 96 & 576 & 485  \\
      \hline
      Source Code Escrow & 59 & 500 & 257 \\
      \hline
      Affiliate License Licensee & 96 & 559 & 475 \\
      \hline
      Post Termination Services & 378 & 461 & 370 \\
      \hline
      Non Transferable License & 237 & 412 & 344\\
      \hline
      Uncapped Liability & 131 & 456 & 406 \\
      \hline
      Irrevocable Or Perpetual License & 128 & 594 & 510 \\
      \hline
      Most Favored Nation & 30 & 455 & 370\\
      \hline
      License Grant & 639 & 431 & 355 \\
      \hline
      Competitive Restriction Exception & 96 & 433 & 361 \\
      \hline
\end{tabular}%
\caption{CUAD: Chosen labels \& tag character lengths}
\label{tab:cuad_datapoints_stats}
\end{table}

In order to assess the performances of both approaches on Long Entities Information Extraction, we based our experiment on the CUAD dataset using only its top 10 labels with the longest entities on average.

This dataset represents a particular business use-case when dealing with contracts or more generally Legal documents, in which entities to extract can be several sentences or even entire paragraphs.

\begin{table}[h]

\begin{tabular}{|c|SSS|SSS|}
\hline
            \textbf{CUAD} & \multicolumn{3}{c|}{\textbf{Token Classification}} & \multicolumn{3}{c|}{\textbf{QA}} \\ \hline
              \textbf{Model} & \textbf{F1} & \textbf{Recall} & \textbf{Precision} & \textbf{F1} & \textbf{Recall} & \textbf{Precision} \\ 
      \hline
      Affiliated License Licensor & 0.00 & 0.00 & 0.00 & 0.00 & 0.00 & 0.00 \\
      \hline
      Source Code Escrow & 0.00 & 0.00 & 0.00 & \textbf{20.00} & 14.29 & 33.33\\
      \hline
      Affiliate License Licensee & 0.00 & 0.00 & 0.00 & \textbf{38.46} & 26.32 & 71.42\\
      \hline
      Post Termination Services & 0.00 & 0.00 & 0.00 & \textbf{39.64} & 30.56 & 56.41\\
      \hline
      Non Transferable License & 0.00 & 0.00 & 0.00 & \textbf{42.55} & 32.79 & 60.61\\
      \hline
      Uncapped Liability & 0.00 & 0.00 & 0.00 & \textbf{39.22} & 27.78 & 66.67\\
      \hline
      Irrevocable Or Perpetual License & 0.00 & 0.00 & 0.00 & \textbf{43.63} & 32.43 & 66.67\\
      \hline
      Most Favored Nation & 0.00 & 0.00 & 0.00 & \textbf{40.00} & 25.00 & 99.99\\
      \hline
      License Grant & 0.00 & 0.00 & 0.00 & \textbf{36.76} & 24.64 & 72.34\\
      \hline
      Competitive Restriction Exception & 0.00 & 0.00 & 0.00 & \textbf{34.04} & 25.81 & 50.00\\
      \hline
      Weighted average & 0.00 & 0.00 & 0.00 & \textbf{37.51} & 27.08 & 63.06\\
      \hline
\end{tabular}%
\caption{Performance of Token Classification and QA LayoutLM information extraction models on CUAD dataset}
\label{tab:benchmark_cuad}
\end{table}

As shown in Tab. \ref{tab:benchmark_cuad}, the token classification approach has a lot of difficulties at capturing long entities whereas the QA approach can predict some of them correctly, even though the F1-score is still low (weighted average F1-score at 37.51).

This could be explained by the fact that in the regular token classification approach, for long documents, since the \textbf{O} tag is predominant among all token classification entities so the model may be focused on reducing that loss overall and therefore predicting no tags.

\subsection{Few-Shot Learning}

In order to assess Few-Shot Learning capabilities, we used the SROIE dataset and we randomly sub-sampled from the documents using different sub-sampling ratios. That setting allows us to recreate an environment where the number of labeled documents is largely reduced.

We decided to take as sub-sampling document ratios \textit{10\%}, \textit{30\%}, \textit{50\%}, \textit{70\%}, \textit{90\%}. This sub-sampling procedure is applied on the training and validation sets but the test set remains the same.

Similarly to the noisy environment setting, we also performed the sub-sampling and training using 5 different random seeds in order to assess the stability and statistical significance of the results.

\begin{figure}[h]
  \centering
  \includegraphics[width = 12cm]{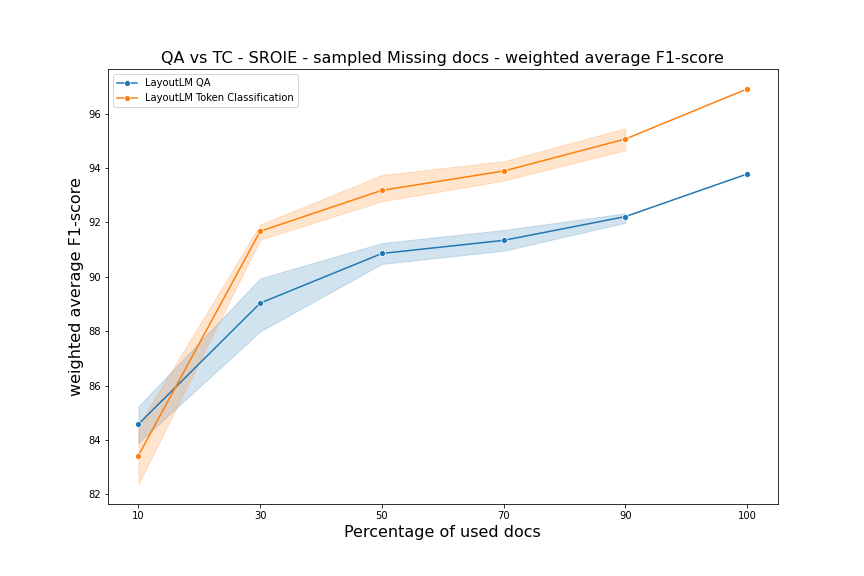}
  \caption{SROIE: QA vs Token Classification performances using partial random missing documents}
  \label{fig:sroie_missing_documents}
\end{figure}

As shown in the performance graph using different sub-sampling ratios in Fig. \ref{fig:sroie_missing_documents}, both approaches are impacted when trained with fewer documents, but still achieving correct results. It is particularly true with the token classification approach which decreased on average from 95.07\% to 83.43\% of weighted F1-score when using only 10\% of documents. On the other hand, the performance drop for the QA approach  was only from 92.76\% to 84.58\%. 

The QA approach seems then slightly more robust to the lack of documents compared to the token classification approach, even though the token classification approach outperforms the QA approach when provided with sufficient documents.

Once again, we notice that the results from the token classification approach have a greater variance than the results from QA.

\subsection{Zero-Shot Learning}

The token classification approach cannot classify unseen classes. We could only assess the Zero-Shot capabilities of the QA approach.

Indeed as mentioned above, we assessed the QA approach using the prepared QA datasets, with potentially generated questions with unseen labels. For example, the model did not necessarily learn entities such as \textit{Header}, \textit{Party} or \textit{Market} during pre-training.

Using the base pre-trained LayoutLM or after fine-tuning on other datasets, we tested the MRC performances on the different datasets.

\begin{table}[h]
\resizebox{\columnwidth}{!}{%
\begin{tabular}{|c|SSS|SSS|SSS|}
\hline
              \textbf{SROIE} & \multicolumn{3}{c|}{\textbf{Pre-trained}} &\multicolumn{3}{c|}{\textbf{FUNSD}} & \multicolumn{3}{c|}{\textbf{Kleister NDA}} \\ \hline
              \textbf{Model} & \textbf{F1}   & \textbf{Precision} & \textbf{Recall}               & \textbf{F1}   & \textbf{Precision} & \textbf{Recall}               & \textbf{F1}   & \textbf{Precision} & \textbf{Recall}   \\ 
      \hline
      Company & 0.00 & 0.00 & 0.00 & \textbf{5.33} & 3.08 & 20.00 & 0.00 & 0.00 & 0.00 \\
      \hline
      Address & 0.00 & 0.00 & 0.00 & 0.00 & 0.00 & 0.00 & 0.00 & 0.00 & 0.00 \\
      \hline
      Total & 0.00 & 0.00 & 0.00 & \textbf{1.54} & 0.86 & 6.82 & 0.00 & 0.00 & 0.00 \\
      \hline
      Date & 54.63 & 86.39 & 39.95 & \textbf{79.67} & 70.70 & 91.25 & 22.32 & 12.59 & 98.11 \\
      \hline
      weighted average & 15.43 & 24.40 & 11.28 &\textbf{24.17} & 20.93 & 32.27 & 6.30 & 3.56 & 27.72 \\
      \hline
\end{tabular}%
}
\caption{Zero-Shot performances of QA LayoutLM information extraction models pre-trained or fine-tuned on different datasets and evaluated on SROIE}
\label{tab:benchmark_zero_shot_sroie}
\end{table}

\begin{table}[h]
\resizebox{\columnwidth}{!}{%
\begin{tabular}{|c|SSS|SSS|SSS|}
\hline
              \textbf{FUNSD} & \multicolumn{3}{c|}{\textbf{Pre-trained}} & \multicolumn{3}{c|}{\textbf{SROIE}} & \multicolumn{3}{c|}{\textbf{Kleister NDA}} \\ \hline
              \textbf{Model} & \textbf{F1}   & \textbf{Precision} & \textbf{Recall}               & \textbf{F1}   & \textbf{Precision} & \textbf{Recall}               & \textbf{F1}   & \textbf{Precision} & \textbf{Recall}   \\ 
      \hline
      Header & 0.00 & 0.00 & 0.00 & \textbf{0.38} & 3.00 & 50.00 & 0.00 & 0.00 & 0.00 \\
      \hline
      Question & 0.00 & 0.00 & 0.00 & 0.00 & 0.00 & 0.00 & 0.00 & 0.00 & 0.00 \\
      \hline
      Answer & 0.00 & 0.00 & 0.00 & 0.00 & 0.00 & 0.00 & 0.00 & 0.00 & 0.00 \\
      \hline
      weighted average & 0.00 & 0.00 & 0.00 & \textbf{0.19} & 0.10 & 2.52 & 0.00 & 0.00 & 0.00 \\
      \hline
\end{tabular}%
}
\caption{Zero-Shot performances of QA LayoutLM information extraction models pre-trained or fine-tuned on different datasets and evaluated on FUNSD}
\label{tab:benchmark_zero_shot_funsd}
\end{table}

\begin{table}[h]
\resizebox{\columnwidth}{!}{%
\begin{tabular}{|c|SSS|SSS|SSS|}
\hline
              \textbf{Kleister NDA} & \multicolumn{3}{c|}{\textbf{Pre-trained}} & \multicolumn{3}{c|}{\textbf{SROIE}} & \multicolumn{3}{c|}{\textbf{FUNSD}} \\ \hline
              \textbf{Model} & \textbf{F1}   & \textbf{Precision} & \textbf{Recall}               & \textbf{F1}   & \textbf{Precision} & \textbf{Recall}               & \textbf{F1}   & \textbf{Precision} & \textbf{Recall}  \\ 
      \hline
      Party & 0.00 & 0.00 & 0.00 & 0.00 & 0.00 & 0.00 & \textbf{1.62} & 0.82 & 57.14 \\
      \hline
      Jurisdiction & 0.00 & 0.00 & 0.00 & 0.00 & 0.00 & 0.00 & \textbf{0.68} & 0.36 & 5.00 \\
      \hline
      Effective Date & 24.71 & 43.75 & 17.21 & 49.26 & 40.98 & 61.73 & \textbf{64.96} & 62.30 & 67.73 \\
      \hline
      Term & 0.00 & 0.00 & 0.00 & 0.00 & 0.00 & 0.00 & 0.00 & 0.00 & 0.00 \\
      \hline
      weighted average & 3.25 & 5.75 & 2.26 & 6.48 & 5.39 & 8.11 & \textbf{9.59} & 8.73 & 40.32 \\
      \hline
\end{tabular}%
}
\caption{Zero-Shot performances of QA LayoutLM information extraction models pre-trained or fine-tuned on different datasets and evaluated on Kleister NDA}
\label{tab:benchmark_zero_shot_kleister_nda}
\end{table}

As shown in the different benchmarks above, the QA approach has only little to none Zero-Shot capabilities depending on the labels it looks for. Especially \textit{Date} labels seems to be easier to detect since it might have seen similar Date labels during fine-tuning. Also it is a relatively standard label since its format has few variants.
Otherwise, predicting totally new labels is not very effective, without any prompt tuning or customization.

%% file: Sections/8-Conclusion.tex
In this paper, we presented an extensive benchmark between the token classification approach and the question answering approach in the Document Information Retrieval context. We showed that the token classification approach is best suited with clean and relatively short entities in terms of both effectiveness and efficiency, whereas the question answering approach can actually be used for this task and represent an credible and robust alternative especially in the cases of noisy datasets or with long entities to extract.

This represents a first study in the literature opposing those approaches on this Key Information Extraction task, that can be completed with further works such as question prompt-tuning or conditional search for the question answering approach, additional experimental settings with nested entities, or even more model architectures that could include image features as shown in the document visual question answering task.